\documentclass[11pt,a4paper]{article}
\usepackage[usenames, dvipsnames]{xcolor}
\usepackage[hyperref]{emnlp2020}
\usepackage{times}
\usepackage{latexsym}
\usepackage{graphicx}
\usepackage{comment}
\usepackage{amsmath}
\usepackage{multirow}

\usepackage{microtype}

\aclfinalcopy % Uncomment this line for the final submission
\DeclareMathOperator*{\argmin}{argmin} % no space, limits underneath in displays
\newcommand{\bing}[1]{\textcolor{ForestGreen}{\textbf{bl}: #1}}

\title{Controllable Text Generation with Focused Variation}

\author{
Lei Shu$^{1*}$ 
, Alexandros Papangelis$^{2}$
, Yi-Chia Wang$^{2}$
, Gokhan Tur$^{2}$,\\
\AND
  Hu Xu$^1$
, Zhaleh Feizollahi$^{2}$
, Bing Liu$^1$
, Piero Molino$^3$ \thanks{Work done while at Uber AI Labs.}
\\
$^1$Department of Computer Science, University of Illinois at Chicago\\
$^2$Uber AI, \\
$^3$Stanford University and ML Collective\\
$^1$shulindt@gmail.com, activebus@gmail.com, liub@uic.edu,\\ 
$^2$al3x.papangelis@gmail.com, yichia.wang@gmail.com, \\
$^2$gokhan.tur@ieee.org, zhaleh.feizollahi@gmail.com\\ 
$^3$pmolino@cs.stanford.edu\\
  \\}

\date{}

\begin{document}
\maketitle
\begin{abstract}

This work introduces Focused-Variation Network (FVN), a novel model to control language generation.
The main problems in previous controlled language generation models range from the difficulty to generate text according to the given attributes, to the lack of diversity of the generated texts.
FVN addresses these issues by learning disjoint discrete latent spaces for each attribute inside codebooks, which allows for both controllability and diversity, while at the same time generating fluent text.
We evaluate FVN on two text generation datasets with annotated content and style, and show state-of-the-art performance as assessed by automatic and human evaluations.

\end{abstract}

\section{Introduction}
\label{introduction}

\begin{table*}[t]
  \centering
  \scriptsize
  %\resizebox{\columnwidth}{!}{
  \begin{tabular}{r|p{0.86\textwidth}}

    CMR & Name[Fitzbillies], EatType[pub], Food[Italian], CustomerRating[decent], Area[Riverside], FamilyFriendly[No], Near[The Sorrento], PriceRange[Moderate] \\
    \hline
    \hline
    
    Text 1 & Fitzbillies is a pub with a decent rating. It is a moderately priced Italian restaurant in riverside near The Sorrento. It is not family-friendly.\\
    \hline
    
    Delex. Text 1 & Name\_SLOT is a EatType\_SLOT with a CustomerRating\_SLOT rating. It is a PriceRange\_SLOT priced Food\_SLOT restaurant in Area\_SLOT near Near\_SLOT. It is FamilyFriendly\_SLOT.\\
    \hline
    
    Agreeable & {\color{red}\underline{Let’s see what we can find on}} {\color{NavyBlue}\textit{Name\_SLOT.}} {\color{red}\underline{I see, well,}} {\color{NavyBlue}\textit{it is an EatType\_SLOT with a CustomerRating\_SLOT rating, also it is a PriceRange\_SLOT priced Food\_SLOT restaurant in Area\_SLOT and near Near\_SLOT, also it is FamilyFriendly\_SLOT,}} {\color{red}\underline{you see?}}\\
    \hline
    
    Disagreeable & {\color{red}\underline{I mean, everybody knows that}} {\color{NavyBlue}\textit{Name\_SLOT is an EatType\_SLOT with a CustomerRating\_SLOT rating. It is a PriceRange\_SLOT priced Food\_SLOT restaurant in Area\_SLOT near Near\_SLOT. It is FamilyFriendly\_SLOT.}}\\
    \hline
    \hline
    
    Delex. Text 2 & Name\_SLOT is a Food\_SLOT place near Near\_SLOT in Area\_SLOT and PriceRange\_SLOT priced. It has a CustomerRating\_SLOT rating. It is an EatType\_SLOT and FamilyFriendly\_SLOT kid friendly.\\
    \hline
    
    Conscientious & {\color{red}\underline{Let’s see what we can find in}} {\color{NavyBlue}\textit{Name\_SLOT.}} {\color{red}\underline{Emm ...}} {\color{NavyBlue}\textit{it is a Food\_SLOT place near Near\_SLOT in Area\_SLOT and PriceRange\_SLOT priced. It has a CustomerRating\_SLOT rating. It is an EatType\_SLOT and FamilyFriendly\_SLOT.}}\\
    \hline
    
    Unconscientious & {\color{red}\underline{Oh god yeah, I don’t know.}} {\color{NavyBlue}\textit{Name\_SLOT is a Food\_SLOT place near Near\_SLOT in Area\_SLOT and PriceRange\_SLOT priced. It has a CustomerRating\_SLOT rating. It is an EatType\_SLOT and FamilyFriendly\_SLOT kid friendly.}}\\

  \end{tabular}
  %}
  \caption{Text generation with focused variations (underlined red denotes personality, italics blue denotes content). The styles are personality traits (dis/agreeable, un/conscientious, extrovert). The content meaning representation and neutral text (Text 1 and 2) are shown at the top. When given a style, the generated text \textit{strictly} follows it. Delex denotes delexicalised text.}
  \label{tab:nlg_example}
  \vspace*{-2ex}
\end{table*}

Recent developments in language modeling~\citep{radford2019language,dai2019transformer,radford2018improving, Holtzman2020TheCC, Khandelwal2020GeneralizationTM} make it possible to generate fluent and mostly coherent text.
Despite the quality of the samples, regular language models cannot be conditioned to generate language %\bing{?? according to?}\piero{according to their definition. If you could condition them, then they would be conditional language models}
depending on attributes.
Conditional language models have been developed to solve this problem, with methods that either train models given predetermined attributes ~\citep{keskar2019ctrl}, use conditional generative models~\cite{kikuchi-etal-2014-single, ficler2017controlling}, fine-tune models using reinforcement learning~\cite{ziegler2019finetuning}, or modify the text on the fly during generation~\cite{Dathathri2020PlugAP}.

As many researchers noted, injecting style into natural language generation can increase the naturalness and human-likeness of text by including pragmatic markers, characteristic of oral language~\cite{biber1991variation, paiva2004framework, mairesse2007personage}.
Text generation with style-variation has been explored as a special case of conditional language generation that aims to map attributes such as the informational content (usually structured data representing meaning like frames with keys and values) and the style (such as personality and politeness) into one of many natural language realisations that conveys them~\cite{novikova2016crowd, novikova2017e2e, Wang18NipsConvAI}.
As the examples in Table~\ref{tab:nlg_example} show, for one given content frame there can be multiple realisations\footnote{We aim to generate delexicalized text containing *\_SLOT tokens, a common practice in a task-oriented dialogue systems.} (e.g., Delex. Texts 1 and 2).
When a style (a personality trait in this case) is injected, the text is adapted to that style (words in red) while conveying the correct informational content (words in blue).
A key challenge is to generate text that respects the specified attributes while at the same time generating diverse outputs, as most existing methods fail to correctly generate text according to given attributes or exhibit a lack of diversity among different samples, leading to dull and repetitive expressions.

Conditional VAEs (CVAE)~\cite{sohn2015learning} and their variants have been adopted for the task and are able to generate diverse texts, but they suffer from posterior collapse and do not strictly follow the given attributes because their latent space is pushed towards being a Gaussian distribution irrespective of the different disjoint attributes, conflating the given content and style.
%\huxu{is it better to give a name of the problem CVAE faced? so you can analyze the problem again in many other places and more about why your solution works well.} \alex{isn't this the `posterior collapse'?}

An ideal model would learn a separate latent space that focuses on each attribute independently.
For this purpose, we introduce a novel natural language generator called Focused-Variation Network~(FVN).
%\footnote{Code for FVN will be made available as open source.}
FVN extends the Vector-Quantised VAE~(VQ-VAE)~\cite{van2017neural}, which is non-conditional, to allow conditioning on attributes (content and style).
Specifically, FVN:
(1) models two disjoint codebooks for content and style respectively that memorize input text variations;
(2) further controls the conveyance of attributes by using content and style specific encoders and decoders;
(3) computes disjoint latent space distributions that are conditional on the content and style respectively, which allows to sample latent representations in a focused way at prediction time.
This choice ultimately helps both attribute conveyance and variability.
As a result, FVN can preserve the diversity found in training examples as opposed to previous methods that tend to cancel out diverse examples.
FVN's disjoint modeling of content and style increases the conveyance of the generated text, while at the same time generating more natural and fluent text.
%Finally, the sampling improves the variation during the prediction phase. 

We tested FVN on two datasets, PersonageNLG \cite{oraby2018controlling} and E2E \cite{dusek.etal2020:csl} that consist of content-utterance pairs with personality labels in the first case, and the experimental results show that it outperforms previous state-of-the-art methods.
A human evaluation further confirms that the naturalness and conveyance of FVN generated text is comparable to ground truth data.

\section{Related Work}
\label{related_work}
Our work is related to CVAE based text generation \cite{bowman2016generating, shen2018improving, zhang2019improve}, where the goal is to control a given attribute of the output text (for example, style) by providing it as additional input to a regular VAE.
%CVAE conditions the generation on both the latent variable and the attribute, assuming that the attribute has a continuous prior distribution. Instead of learning the full latent space over all values of the given attribute, it models a distribution for each value, over the full latent space. %The boundaries of each class in CVAE are not sharp. 
%Once there are multiple attributes, it assumes the prior distribution given the joint of attributes. This methodology is not easy to extend with multiple disentangle attributes.  
For instance, the controlled text generation method proposed by \citet{hu2017toward} extends VAE and focuses on controlling attributes of the generated text like sentiment and style.
Differently from ours, this method does not focus on generating text from content meaning representation (CMR) or on diversity of the generated text.
\cite{song2019exploiting} use a memory augmented CVAE to control for persona, but with no control over the content.
%CVAE-based text generation models the style's or topic's variation, but they do not aim at the fidelity to the meaning representation of the text as well. 

The works of \cite{oraby2018controlling, harrison2019maximizing, oraby2019curate} on style-variation generators adopt sequence-to-sequence based models and use human-engineered features~\cite{juraska2018characterizing} (e.g. personality parameters or syntax features) as extra inputs alongside the content and style to control the generation and enhance text variation. 
However, using human-engineered features is labor-intensive and, as it is not possible to consider all possible feature combinations, performance can be sub-optimal.
In our work we instead rely on codebooks to memorize textual variations.

%Our work is also related to discrete representation learning.
%We adopt the Vector-Quantised VAE (VQ-VAE)\cite{van2017neural},
%a simple yet powerful generative model that learns such discrete representations. It differs from VAEs in two key ways: the VQ-VAE encoder network outputs discrete, rather than continuous, codes; and the prior is learned dynamically rather than being static. Using the vector quantization (VQ) method allows the model to circumvent issues of ``posterior collapse”, where the latent features are ignored when they are paired with a powerful autoregressive decoder, which is typically observed in the VAE framework. Although VQ-VAE is a general VAE model, it does not focus on generating text from meaning or style representation. 
%We use VQ-VAE to learn and memorize the latent-features of semantics and style from training examples. 

There is a variety of works that address the problem of incorporating knowledge or structured data into the generated text (for example, entities retrieved from a knowledge base) \cite{ye2020variational}, or that try generate text that is in line with some given story \cite{rashkin2020plotmachines}.
None of these works focuses specifically on generating text that conveys content while at the same time controlling style.
Last, there are works such as \cite{rashkin2018towards} that focus on generating text consistent with an emotion (aiming to create an empathetic agent) without, however, directly controlling the content.

\section{Methodology}
\label{methodology}

\begin{figure*} %  figure placement: here, top, bottom, or page
    \centering
    \includegraphics[width=0.9\textwidth]{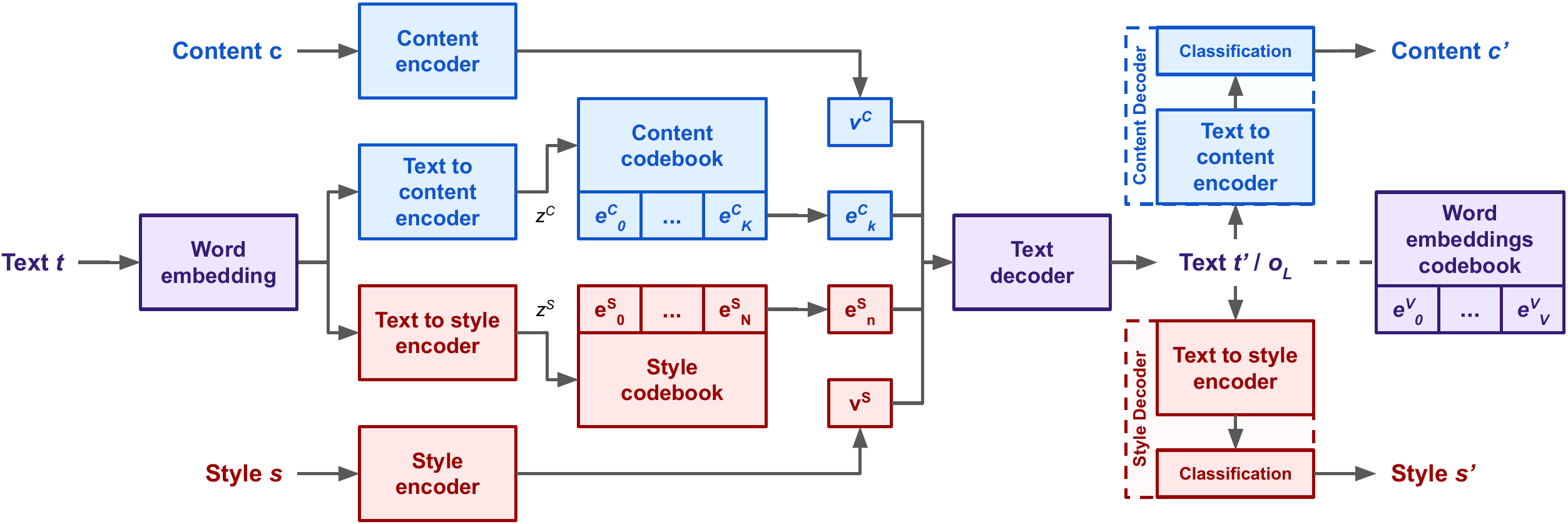}
    \caption{Focused-Variation Network (FVN) has four encoders (text-to-content encoder, text-to-style encoder, content encoder and style encoder), two codebooks (for content and style), and one text decoder. The training data contains ground-truth text with associated content and style. The text decoder uses $v^C$ and $v^S$, latent vectors of content and style, as well as the latent vectors $e^C_k$ and $e^S_n$ from codebooks (the nearest to the $z^C$ and $z^S$ vectors produced by the text-to-content and text-to-style encoders) to generate text back. To further control the content and style of the generated text, we feed the $o_L$ output vectors of the generated text $t'$ to text encoders (content and style). $o_L$ are aligned to a word embedding codebook.}
    \label{fig:fvn}
    \vspace{-3mm}
\end{figure*}    

Our proposed FVN architecture (Figure~\ref{fig:fvn}) has the goal to generate diverse texts that respect every attribute provided as controlling factor.
We describe a specific instantiation where the attributes are content (a frame CMR containing slots keys and values) and style (personality traits).
However, the same architecture can be used with additional attributes and / or with different types of content attributes (structured data tables and knowledge graphs for instance) and style attributes (linguistic register, readability, and many others).
%\huxu{probably mention the problem of CVAE again and summarize why your solution may work.}
To encourage conveyance of the generated texts, FVN learns disjoint discrete content- and style-focused representation codebooks inspired by VQ-VAE as extra information along with the representations of intended content and style, which avoids the posterior collapse problem of VAEs.

%\huxu{the following two paragraphs seem long if they aim to give the high-level concepts of training and testing.} \huxu{missing reference to figure 1?}
During training, FVN receives as input an intended content $c$ and style $s$ as well as a reference text $t$.
The reference text is passed through two encoders (text-to-content and text-to-style), while content and style are encoded with a content encoder and a style encoder.
The text-to-content encoder maps input text $t$ into a content latent vector $z^C$ and the text-to-style encoder maps the input text $t$ into a latent style vector $z^S$.
The closest vectors to $z^C$ and $z^S$ from the content codebook $e^C$ and style codebook $e^S$, $e^C_k$ and $e^S_n$, are selected.
%The codebooks learned during training memorize the latent vectors of the text-to-content and text-to-style encoders.
The content encoder encodes the intended content frame into a latent vector $v^C$ and the style encoder encodes the intended style into a latent vector $v^S$ .
A text decoder then receives $e^C_k$, $e^S_n$, $v^C$ and $v^S$ and generates the output text $t'$.
The generated text is subsequently fed to a content and a style decoder that predict the intended content and style.
%These backward prediction tasks force the generators to capture the nuances of style and content and thus better control these aspects in the generated text.

At prediction time (Figure \ref{fig:fvn_content_style_decoders_prediction}), only content $c$ and style $s$ are given, and in order to obtain $e^C_k$ and $e^S_n$ without an input text, we A) collect a distribution over the codebook indices by counting, for each training datapoint containing a specific value for $c$ and $s$, the amount of times a specific index is used, and B) sample $e^C_k$ and $e^S_n$ from these frequency distributions.
These disjoint distributions allow the model to focus on specific content and style by using them for conditioning and the sampling allows for variation, hence the name of focused variation.
$v^C$ and $v^S$ obtained from the content and style encoders and the sampled $e^C_k$ and $e^S_n$ are provided to the text generator that generates $t'$.
%In order to do so, we construct two probability distributions $P(K|C)$ and $P(N|S)$ over the indices of $e^C$ and $e^S$, given content and style respectively.
%In order to construct those distributions, training text data is fed into the trained text-to-content and text-to-style encoders to obtain $z^S$ and $z^C$.
%The indices $k$ and $n$ of the nearest vectors $\argmin_k \|e^C_k - z^C\|_2$ and $\argmin_n \|e^S_n - z^S\|_2$ in the codebooks $e^C$ and $e^S$ are collected.
%The indices associated with each content $c \in C$ and style $s \in S$ are counted and normalized to obtain the probability distributions $P(K|C)$ and $P(N|S)$ over the indices $k \in e^C$ and $n \in e^S$ respectively.

%At prediction time (Figure \ref{fig:fvn_content_style_decoders_prediction}), given the intended content $c$ and the intended style $s$ we obtain $v^C$ and $v^S$ using the content encoder and the style encoder respectively.
%To obtain $e^C_k$ and $e^S_n$ we sample $k$ and $n$ from $P(K|c)$ and $P(N|s)$ and select the $k$-th and $n$-th element from the $e^C$ content codebook and $e^S$ style codebook respectively. 
%We provide the generated $v^C$ and $v^S$ together with the sampled $e^C_k$ and $e^S_n$ to the text decoder to generate new text.

The rest of this section will detail each component and the training and prediction processes.

\subsection{Encoding and Codebooks}
\label{sec:encoding_codebooks}

As shown in Figure \ref{fig:fvn}, FVN uses four encoders and one decoder during training: the text-to-content encoder $\text{Enc}^\textit{TC}(\cdot)$, the text-to-style encoder $\text{Enc}^\textit{TS}(\cdot)$, the content encoder $\text{Enc}^{C}(\cdot)$, the style encoder $\text{Enc}^{S}(\cdot)$, and the text decoder $\text{Dec}(\cdot)$.

\textbf{Text-to-* encoders}
The text-to-content encoder $\text{Enc}^\textit{TC}(\cdot)$ encodes a text $t$ to a dense representation $z^C \in \mathcal{R}^D$ while the text-to-style encoder $\text{Enc}^\textit{TS} ( \cdot )$ encodes a text $t$ to a dense representation $z^S \in \mathcal{R}^D$: $z^C = \text{Enc}^\textit{TC} (t)$ and $z^S = \text{Enc}^\textit{TS} (t)$.

%\begin{align}
%    z^C &= \text{Enc}^\textit{TC} (t), \\
%    z^S &= \text{Enc}^\textit{TS} (t).
%\label{eq:textenc}
%\end{align}

In order to learn disjoint latent spaces for the different attributes we want to model, we train two codebooks, one for content $e^C \in \mathcal{R}^{K \times D}$ and one for style $e^S \in \mathcal{R}^{N \times D}$.
They are shown as $[e^C_1, \dots, e^C_K]$ and $[e^S_1, ..., e^S_N]$ in Figure~\ref{fig:fvn}.

These two codebooks are used to memorize the latent vectors for text-to-content variation and text-to-style variation learned during training.
Instead of using the $z^C$ and $z^S$ vectors as inputs to the decoder, we find their nearest latent vectors in the codebooks $e^C_k$ and $e^S_n$ and use those nearest latent vectors for decoding the text instead of the original encoded dense representation.
Formally, $k = \argmin_i \|z^C - e^C_i\|_2$ and $n = \argmin_j \|z^S - e^S_j\|_2$.

%\begin{align}
%    k &= \argmin_i \|z^C - e^C_i\|_2, \\
%    n &= \argmin_j \|z^S - e^S_j\|_2.
%\label{eq:nearest}
%\end{align}

Like in VQ-VAE, we use the $l$2-norm error to move the latent vectors in the codebooks $e$ towards the same space of the encoder outputs $z$:
\begin{small}
\begin{align}
    \mathcal{L}_{\textit{VQ}}^C &= \|sg(z^C) - e^C_k \|^2_2 + \beta^C\|z^C - sg(e^C_k)\|^2_2,  \\
    \mathcal{L}_{\textit{VQ}}^S &= \|sg(z^S) - e^S_n \|^2_2 + \beta^S\|z^S - sg(e^S_n)\|^2_2,
\label{eq:vqloss}
\end{align}
\end{small}
where $sg(\cdot)$ stands for the stop gradient operator.
%Considering the volume of the embedding space, it can grow arbitrarily if the latent vector $e$ does not train as fast as the encoder parameters. To make sure the encoder commits to a latent vector and its output does not grow, add a commiTCent loss,

\textbf{Style and content encoders}
The content encoder encodes a CMR $c$ treating it as a sequence of tokens and producing a matrix $V^C \in \mathcal{R}^{L' \times D}$, where $L'$ is the length of $c$, from which the last element $v^C \in \mathcal{R}^D$ is returned.
The style encoder encodes a style $s$ and obtains a dense representation $v^S \in \mathcal{R}^D$ selecting the last element of the matrix $V^S \in \mathcal{R}^{L'' \times D}$.
Ultimately, $v^C = \text{Enc}^{M} (m)$ and $v^S = \text{Enc}^{S} (s)$.

%\begin{align}
%    v^C &= \text{Enc}^{M} (m) , \\
%    v^S &= \text{Enc}^{S} (s) .
%\label{eq:enc}
%\end{align}
%The latent codebook vectors $e^C_k$ and $e^S_n$ contain both text-variation information and the meaning/style information (after controlling). The reason for using both latent vectors and encoded vectors ($v^C$ and $v^S$) is that there can still be errors about meaning/style information in the latent vectors. The encoded vectors ($v^C$ and $v^S$) give the decoder a correct sense of meaning and style.
%\huxu{seems this paragraph jumped from nowhere? missing some sentence to draw a context.}
Both sets of vectors, $e$ and $v$ are needed as the former learn to memorize the encoded inputs $z$, while the latter learn regularities in the attributes.

\subsection{Text Decoder}
\label{sec:text_decoder}
%\huxu{better to say both inputs and outputs to form the goal of decoder.}
The decoder takes the $e^C_k$, $e^S_n$, $v^C$ and $v^S$, which encode content and style, as input and decodes text $t'$.
We use an LSTM network to model our decoder and provide the initial hidden state $h_0$ and initial cell state $c_0$.
The initial hidden state is the concatenation of $e^C_k$ and $e^S_n$, while the initial cell state is the concatenation of $v^C$ and $v^S$: $c_0 = v^C \circ v^S$ and $h_0 = e^C_k \circ e^S_n$.

%\begin{align}
%    c_0 &= v^C \circ v^S, \\
%    h_0 &= e^C_k \circ e^S_n.
%\label{eq:initialdecoder}
%\end{align}

When we decode the $l$-th word, we encode the previous word $t'_{l-1}$ and pay attention to the encoded sequence of content $v^C$ and style $v^S$ using the last hidden state as a query.
Since both content and style are sequences of words, the attention mechanism can help figure out which part of them is important for decoding the current word.
We concatenate the embedded previous output word and the attention output as the input for LSTM $x_l$.
The LSTM updates the hidden state, cell state and produces an output vector $g_l \in \mathcal{R}^{2D}$.
Since we want to feed the generated text back to text encoders for additional control, we reduce $g_l$ to a word embedding dimension vector $o_l$ by a linear transformation.
Finally, we map $o_l$ to the size of the vocabulary and apply $\text{softmax}$ to obtain a probability distribution over the vocabulary. \vspace{-4mm}

\begin{small}
\begin{align}
    x_l &= \text{Emb} (t'_{l-1}) \circ \text{Attn} (h_{l-1}, V^C \circ V^S),\\
    g_l, (h_l, c_l) &= \text{LSTM} \Big(x_l, (h_{l-1}, c_{l-1})\Big),\\
    o_l &= W_\textit{emb}\cdot g_l + b_\textit{emb},\\
    P(t'_l) &= \text{softmax} (W_V \cdot o_l + b_V).
\label{eq:decoder}
\end{align}
\end{small}

\vspace{-4mm}
The loss for text decoding is the sum of cross entropy loss of each word os $\mathcal{L}_\textit{Dec} = -\sum_l \log P(t'_l)$.

%\begin{align}
%    \mathcal{L}_\textit{Dec} &= -\sum_l \log P(t'_l).
%\label{eq:decloss}
%\end{align}

\subsection{Content and Style Decoders}
\label{sec:content_style_decoders}

%\begin{figure} %  figure placement: here, top, bottom, or page
%    \centering
%    \includegraphics[width=0.7\columnwidth]{fvn_content_style_decoders.pdf}
%    \caption{FVN architecture from text decoding to content and style decoding. To further control the content and style of the generated text, we feed the $o_L$ output vectors of the generated text $t'$ to text encoders (content and style). $o_L$ are aligned to a word embedding codebook.}
%    \label{fig:fvn_content_style_decoders}
%\end{figure}

To ensure the generated text $t'$ conveys the correct content and style, 
we feed them to content and style decoders to perform backward prediction tasks that better control the generator.
The decoders contain two components: we first reuse the text-to-content and text-to-style encoders to encode the embedded predicted text $o_L$ and obtain latent representations $z'^C$ and $z'^S$, and then we classify them to predict content $c'$ and style $s'$, as shown in the right side of Figure~\ref{fig:fvn}: $z'^C = \text{Enc}^{\textit{TC}} (o_L)$ and $z'^S = \text{Enc}^{\textit{TS}} (o_L)$.
%\begin{align}
%    z'^C &= \text{Enc}^{\textit{TC}} (o_L) , \\
%    z'^S &= \text{Enc}^{\textit{TS}} (o_L) .
%\label{eq:decodedtextenc}
%\end{align}
$\text{Enc}^\textit{TC}(\cdot)$ and $\text{Enc}^\textit{TS}(\cdot)$ denote the same text-to-content and text-to-style encoders we defined previously.
This design is inspired by work on text style transfer~\cite{dossantos18fightingoffensivelanguage}.

%Controlling modules can enhance the MR's correctness and style's correctness in text generation. Considering our proposed methods use the latent vectors $e^C_k$ and $e^S_n$ for decoding the text, we apply controlling on the latent vectors as well as encoded vectors from the generated text.

Both $z'$ vectors and $e$ vectors are used by two classification heads $F^C$ (multi-label) and $F^S$ (multi-class) for predicting content and style respectively in order to force those vectors to encode attribute information.
We use $g$ to denote the $g$-th element in the set of possible key-value pairs in the CMR and $m(\cdot)$ to represent an indicator function that returns whether the $g$-th element is in the ground-truth CMR. \vspace{-4mm}

\begin{small}
\begin{align}
    P\Big(y^C_z(g) = m(g)\Big) &= F^C (z'^C) ,\\
    P(y^S_z = s) &= F^S (z'^S) ,\\
    P\Big(y^C_e(g) = m(g)\Big) &= F^C (e^C_k) , \\
    P(y^S_e = s) &= F^S (e^S_n) .
\label{eq:classification}
\end{align}
\end{small} \\
The loss for training the two prediction heads is:
\begin{scriptsize}
\begin{align}
    \mathcal{L}_\textit{CTRL}& =
     - \sum_g\log P\Big(y^C_e(g) = m(g)\Big) - \log P(y^S_e = s) \nonumber \\
    & -\sum_g\log P\Big(y^C_z(g) = m(g)\Big) - \log P(y^S_z = s) . \\\nonumber
\label{eq:classificationloss}
\vspace{-3mm}
\end{align}
\end{scriptsize}

\vspace{-10mm}

Finally, we also adopt vector quantization by mapping each generated word's representation $o_l$ to the word embedding $e^V \in \mathcal{R}^{|V| \times D}$ to map the output of the decoder and the input of text encoders in the same space.
This is needed because the text-to-* encoders expect as input text embedded using word embeddings, but in this case we are providing $o_L$ as input, and without this vector quantization loss, $o_L$ will not be in the same space of the embeddings.
As a result, there is another VQ loss: $\mathcal{L}_\textit{VQ}^V = \|sg(o_l) - e_v^V \|^2_2 + \beta^V\|o_l - sg(e_v^V)\|^2_2$.

%\begin{align}
%    \mathcal{L}_\textit{VQ}^V &= &\|sg(o_l) - e_v^V %\|^2_2 + \beta^V\|o_l - sg(e_v^V)\|^2_2 .
%\label{eq:embloss}
%\end{align}

The total loss minimized during training is the sum of the losses for decoding the text, predicting the content and style, the VQ-loss from two codebooks, and the VQ-loss for word embedding: $\mathcal{L} = \mathcal{L}_\textit{Dec} + \mathcal{L}_\textit{CTRL} + \mathcal{L}_\textit{VQ}^C + \mathcal{L}_\textit{VQ}^S + \mathcal{L}_\textit{VQ}^V$.

%\begin{align}
%\mathcal{L} = \mathcal{L}_\textit{Dec} + %\mathcal{L}_\textit{CTRL} + \mathcal{L}_\textit{VQ}^C + %\mathcal{L}_\textit{VQ}^S + \mathcal{L}_\textit{VQ}^V  .
%\label{eq:loss}
%\end{align}

\subsection{Prediction}

\begin{figure} %  figure placement: here, top, bottom, or page
    \centering
    \includegraphics[width=\columnwidth]{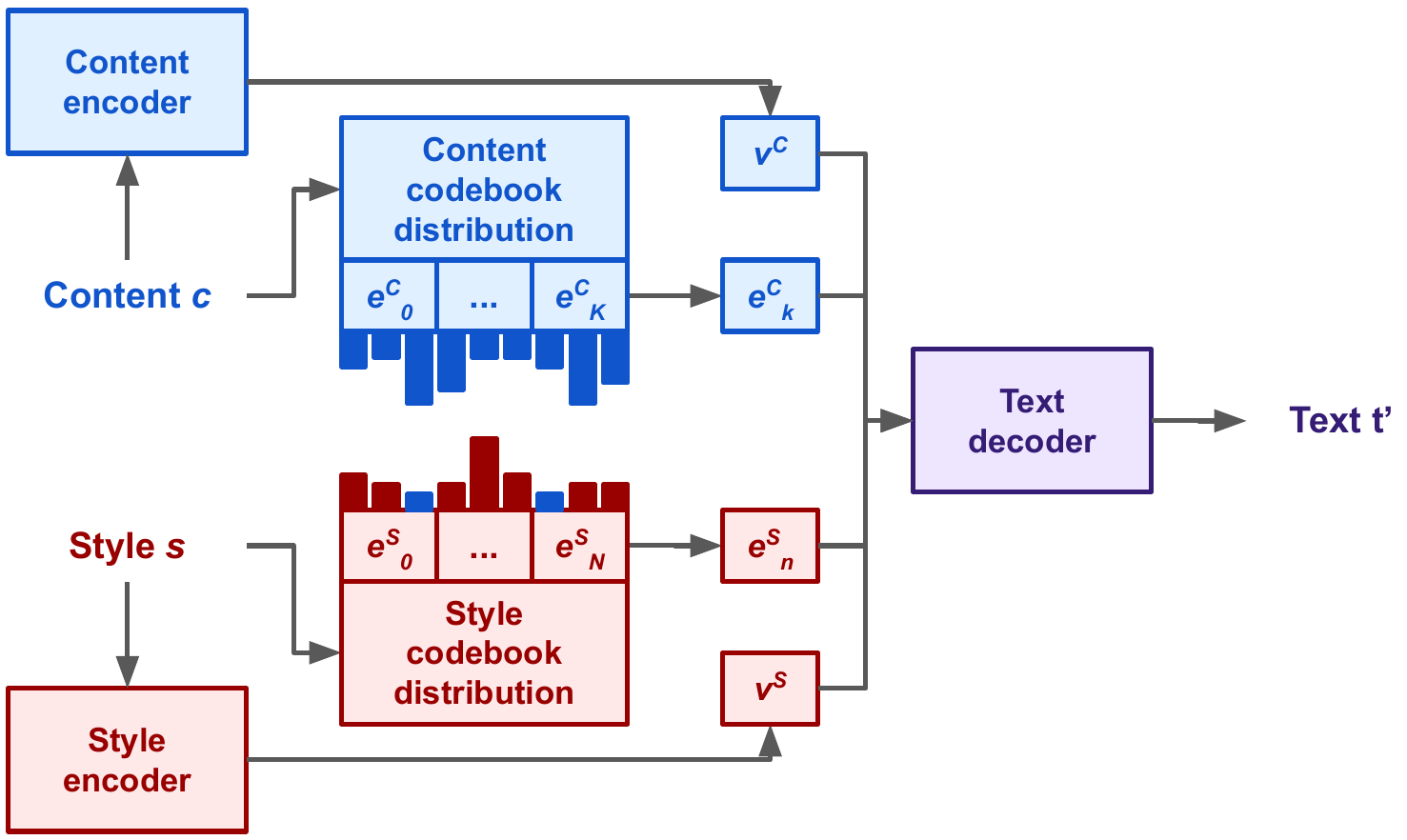}
    \caption{At prediction time we encode $c$ and $d$ with encoders to obtain $v^C$ and $v^S$ and we select $e^C_k$ by sampling $k \sim P(K|C=c)$ and $e^S_n$ by sampling $n \sim P(N|S=s)$. Those four vectors are provided as input to the text decoder to generate text.}
    \label{fig:fvn_content_style_decoders_prediction}
\end{figure}

The whole prediction process is depicted in Figure \ref{fig:fvn_content_style_decoders_prediction}.
The trained text decoder expect four inputs: $v^C$, $v^S$, $e^C_k$, and $e^S_n$.
At prediction time, only content $c$ and style $s$ are given.
We can obtain $v^C$, $v^S$ by providing $c$ and $s$ to their respective encoder, but we also need to obtain $e^C_k$ and $e^S_n$ without input text.
At the end of the training phase, we map each content $c \in C$ and style $s \in S$ to the indices in the $e^C$ and $e^S$ codebooks by first obtaining $z^S$ and $z^C$ vectors from the training data associated with $c$ and $s$, we find the index of the closest codebooks vectors by $\argmin_k \|e^C_k - z^C\|_2$ and $\argmin_n \|e^S_n - z^S\|_2$ and count how many times each index $k \in K$ was the closes to each $c \in C$ and likewise for indices $n \in N$ for each $s \in S$.
By normalizing the counts, we obtain a distribution $P(K|C)$ for content and a distribution $P(N|S)$ for style. 
The construction of the two distributions is performed only once at the end of the training phase.

To obtain $e^C_k$ at prediction time, we select the $k$ vector of the codebook by sampling $k \sim P(K|C=c)$ and likewise to obtain $e^S_n$ with $n \sim P(N|S=s)$.
Sampling from those distributions allows to both focus on specific content and style disjointly by conditioning on them, while at the same time allowing variability because of the sampling (we refer to this procedure as focused variation).
$v^C$, $v^S$, $e^C_k$, and $e^S_n$ are finally provided as inputs to the decoder to generate the text $t'$.
Content and style decoders mentioned in the training section are not needed for prediction.
%\huxu{might need more words on discussion of design choices? like why something is not adopted if the reviewer may imagine ?}

\section{Experiments}
To test the capability of FVN to generate diverse texts that convey the content while adopting a certain style, we use the PersonageNLG text generation dataset for dialogue systems that contains CMR and style annotations.
To test if FVN can convey the content (both slots and values) correctly on an open vocabulary, with complex syntactic structures and diverse discourse phenomena, we use the End-2-End Challenge dataset (E2E), a text generation dataset for dialogue systems that is annotated with CMR.

\subsection{Datasets and Baselines}
%\liu{it is good to organize this section otherwise it is hard to read and to find things} 
%\huxu{put the two purposes of these datasets into a single paragraph between experiment section and the first subsection? people don't know why you want to use two datasets.}

\textbf{PersonageNLG} contains 88,855 training and 1,390 test examples.
We reserve 10\% of the train set for validation.
There are 8 slots in the CMR and 5 kinds of style: agreeable, disagreeable conscientious, unconscientious, and extravert personality traits.
The styles are evenly distributed in both train and test sets.
All slots' values are delexicalized.
We model the focused variation distribution of the content by jointly modeling the presence of slot names in the CMR, e.g. $P(K|PriceRange \in c \text{ }and\text{ } FoodType \in c)$, because there are no slot values.
Style is modeled as a single categorical variable, e.g. $P(N|s=Agreable)$.

\textbf{E2E} contains 42,061 training examples (4,862 CMRs), 4,672 development examples (547 CMRs) and 4,693 test examples (630 CMRs).
Like in the PersonageNLG dataset, there are 8 slots in the CMR.
%See Table \ref{tab:dataseTCr_e2e} for more details.
Each CMR has up to 5 realisations (references) in natural language.
Differently from PersonageNLG, the CMRs in the different splits are disjoint and the texts are lexicalized.
Following the challenge guidelines~\cite{duvsek2018findings}, we delexicalized only `name' and `near' keeping the remaining slots' values.
Since the E2E dataset does not have style annotations but has lexicalized texts, we model the CMR in the same way we did for PersonageNLG, but we replace the style codebook with a slot-value codebook that help the text decoder generating the slot values in the CMR.
We build the focused variation distribution for every slot-value independently over the codebook indices, e.g. $P(N|s=PriceRange[high])P(N|s=FoodType[French])...$.
During prediction we sample codes for each slot value in the CMR and use their average to condition text decoding.
This is particularly useful when the surface forms in the output text are not the slot values themselves, e.g. when ``PriceRange[high]'' should be generated as ``expensive'' rather than ``high''.

In both experiments we use $K=512$ and $N=|V|$ as codebooks sizes.
More dataset details are shown in Appendix~\ref{app:experiment-details} Table~\ref{tab:dataseTCr} and Table~\ref{tab:dataseTCr_e2e}, while details about preprocessing, architecture, and training parameters are provided in Appendix~\ref{app:implementation_details}.

%\huxu{the following paragraph is too long? probably make a small table, columned by name, purpose, misc. setting, etc.?}

We compare our proposed model against the best performing models in both datasets.
All of them are sequence-to-sequence based models.
For PersonageNLG, \textbf{TOKEN}, \textbf{CONTEXT} (from \cite{oraby2018controlling}) are variants of the TGEN~\cite{novikova2017e2e} architecture, while \textbf{token-m*} and \textbf{context-m*} are from  \cite{harrison2019maximizing} (which adopt OpenNMT-py~\cite{opennmt} as the basic encoder-decoder architecture).
token-* baselines use a special style token to provide style information while context-* baselines use 36 human defined pragmatic and aggregation-based features to provide style information
`-m*' indicates variants of how the style information is injected into the encoder and the decoder.
For the E2E challenge dataset, \textbf{TGEN}~\cite{novikova2017e2e}, \textbf{SLUG}~\cite{juraska2018deep}, and \textbf{Thomson Reuters NLG}~\cite{davoodi2018e2e, smiley2018e2e} are the best performing models.
They have different architectures, re-rankers, beam search and data augmentation strategies.
More details are provided in Appendix~\ref{app:baseline-details}.

The results of the baselines~\cite{oraby2018controlling,harrison2019maximizing} are taken from their original papers, but it's unclear if they were evaluated using a single or multiple references (for this reason they are marked with $\dagger$), but since these models are not dependent on sampling from a latent space, we would not expect that to change performance.
%we compare our proposed model and variants against the best performing models in the challenge: TGEN~\cite{novikova2017e2e}, SLUG~\cite{juraska2018deep}, and Thomson Reuters NLG~\cite{davoodi2018e2e, smiley2018e2e}.

We also compare to conditional VAEs:
\textbf{CVAE} implements the conditional VAE \cite{sohn2015learning} framework.
\textbf{Controlled CVAE} implements the controlled text generation \cite{hu2017toward} framework.
The architecture and hyper-parameters of CVAE and controlled CVAE are the same as FVN.

The FVN ablations used in our evaluation are:
(1) \textbf{FVN-ED}
%\lei{should we change it to seq2seq incase reviewer try to locate seq2seq as baseline?}
does not use the codebooks, only uses the content and style encoders and decoders, and is equivalent to an attention-augmented sequence-to-sequence model;
(2) \textbf{FVN-VQ} does not use the content and style encoders and decoders, it directly uses the sampled latent vector for text decoding (\ref{sec:content_style_decoders});
(3) \textbf{FVN-EVQ} does not use content and style decoders;
(4) \textbf{FVN} is the full network.
Refer to Table~\ref{tab:modules} for architecture details. 
All VAEs and FVN variants are evaluated using multiple references because the sampling from latent space may lead to generate a valid and fluent text that n-gram overlap metrics would not score high when evaluated against a single reference.

\subsection{Automatic Evaluation}

\begin{table}
  \centering
  \resizebox{\columnwidth}{!}{
  \begin{tabular}{r|c|c|c|c}
    Model                                   &BLEU       &NIST       &METEOR     &ROUGE-L\\
    \hline
    \hline
    %NOSUP$^\dagger$          &0.2774     &4.2859     &0.3488     &0.4567 \\
    TOKEN$^\dagger$          &0.3464     &4.9285     &0.3648     &0.5016 \\
    CONTEXT$^\dagger$        &0.3766     &5.3437     &0.3964     &0.5255 \\
    token-m1$^\dagger$       &0.4904     &-          &-          &-      \\
    token-m2$^\dagger$       &0.4810     &-          &-          &-      \\ 
    token-m3$^\dagger$       &0.4906     &-          &-          &-      \\ 
    context-m1$^\dagger$     &0.5530     &-          &-          &-      \\ 
    context-m2$^\dagger$     &0.5229     &-          &-          &-      \\ 
    context-m3$^\dagger$     &0.5598     &-          &-          &-      \\ 
    \hline   
    CVAE                     &0.9        &9.766      &0.449      &0.702 \\
    Controlled CVAE          &0.928      &9.957      &0.463      &0.721 \\
    \hline   
    FVN-ED                   &0.802      &7.872      &0.378      &0.696\\
    FVN-VQ                   & 0.887	&8.985	&0.423	&0.715\\%&0.782      &4.45       &0.374      &0.692 \\
    FVN-EVQ                  & 0.94	&\textbf{10.129}	&0.476	&0.748 \\%&0.939      &9.897      &\textbf{0.479} &\textbf{0.764} \\
    FVN                      & \textbf{0.965}	&9.946	&\textbf{0.486}	&\textbf{0.768}\\%&\textbf{0.948} &\textbf{10.103} &0.478 &0.762 \\  
  \end{tabular}
  }
  \caption{Quality Evaluation for PersonageNLG.}
  \label{tab:quality}
\end{table}

\begin{table}
  \centering
  \resizebox{0.9\columnwidth}{!}{
  \begin{tabular}{r|c|c|c}
    Model           &Precision      &Recall         &$F_1$ score     \\
    \hline
    \hline   
    CVAE            &0.961          &0.942          &0.952\\
    Controlled CVAE &0.961          &0.969          &0.965\\
    \hline
    FVN-ED           &\textbf{0.997}	        &0.748	        &0.855\\
    FVN-VQ         	& 0.87	&0.799	&0.833\\
    FVN-EVQ     &0.963          &0.989          &0.976\\
    FVN            & 0.987	&\textbf{1.0}	&\textbf{0.994}\\
  \end{tabular}
  }
  \caption{Content Correctness Evaluation for PeronageNLG. Micro precision, recall and $F_1$ score for ``*\_SLOT" tokens.}
  \label{tab:content_correctness}
\end{table}

\begin{table}
  \centering
  \resizebox{0.9\columnwidth}{!}{
  \begin{tabular}{r|c|c|c}
    Model           &Precision       &Recall       &$F_1$ score     \\
    \hline
    \hline   
    CVAE 	        &0.973	        &0.973	        &0.973\\
    Controlled CVAE & 0.981	        &0.981	        &0.981\\\hline
    FVN-ED           &0.996          &0.996          &0.996\\
    FVN-VQ 	        & 1.0     	&1.0	        &1.0\\
    FVN-EVQ 	& 1.0     	&1.0	        &1.0\\
    FVN 	        &\textbf{1.0}	&\textbf{1.0}	&\textbf{1.0}\\
  \end{tabular}
  }
  \caption{Style Evaluation on PersonageNLG. Macro precision, recall and $F_1$ score for the style of generated text based on a separately trained style classifier.}
  \label{tab:style}
\end{table}

\begin{table}
\resizebox{0.9\columnwidth}{!}{
\begin{tabular}{r|c|c|c|c}
 Model           &1-gram       &2-gram       &3-gram    &4-gram     \\
 \hline
 \hline
ground truth 	& 0.74	&0.902	&0.924	&0.905 \\
\hline
CVAE 	& \textbf{0.738}	&\textbf{0.896}	&\textbf{0.919}	&0.902\\
Controlled CVAE 	& 0.715	&0.869	&0.902	&0.899\\
\hline   
FVN-ED 	& 0.508	&0.618	&0.668	&0.71\\
FVN-VQ 	& 0.68	&0.849	&0.896	&0.894\\%	& \textbf{0.836}	&\textbf{0.922}	&0.903	&0.862\\
FVN-EVQ 	& \textbf{0.738}	&0.883	&0.907	&0.901\\
FVN 	& 0.720	&0.870	&0.906	&\textbf{0.904}\\
  \end{tabular}
  }
  \caption{Diversity Evaluation on PersonageNLG. Distinct n-grams between generated texts and ground truth.}
  \label{tab:diversity}
\end{table}

\begin{table}
  \centering
  \resizebox{0.9\columnwidth}{!}{
  \begin{tabular}{r|c|c|c|c}
    Model           &BLEU       &NIST       &METEOR     &ROUGE-L\\
    \hline
    \hline
    TGEN            &0.659 &8.609 &0.448 &0.685 \\
    SLUG            &0.662 &8.613 &0.445 &0.677 \\
    Thomson Reuters NLG&0.681	&8.778 &	0.446&	0.693\\
    Thomson Reuters NLG&0.674	&8.659	&0.450&	0.698\\
    \hline
    CVAE 	& 0.377	&6.624	&0.336	&0.525\\
    Controlled CVAE 	& 0.404	&6.852	&0.346	&0.544\\
    \hline
    FVN-ED 	& 0.665	&8.359	&0.428	&0.699\\
    FVN-VQ 	& 0.681	&8.864	&0.422	&0.698\\
    %FVN-EVQ & 0.692	&8.869	&0.437	&\textbf{0.714} \\%& 0.672	&8.685	&0.424	&0.698\\
    %FVN 	& \textbf{0.69}	&\textbf{8.895}	&0.421	&\textbf{0.704}\\
    
    %FVN 	& \textbf{0.707}	&\textbf{9.032}	&\textbf{0.452}	&0.708 %& \textbf{0.691}	&\textbf{8.847}	&\textbf{0.451}	&\textbf{0.707}
    %\\\hline
    FVN-EVQ 	& 0.711	&\textbf{9.066}	&\textbf{0.453}	&\textbf{0.721} 	\\
FVN 	& \textbf{0.714}	&9.004	&0.451	&0.719 	\\\hline
   
  \end{tabular}
  }
  \caption{Quality Evaluation on E2E.
  %Following the challenge guidelines, our systems use partial delexicalisation (with `name' and `near' attributes replaced by placeholders during generation)
  }
  \label{tab:e2e}
\end{table}

\begin{table}

\resizebox{0.9\columnwidth}{!}{
\begin{tabular}{r|c|c|c|c}
\multicolumn{5}{c}{}\\
 Model           &1-gram       &2-gram       &3-gram    &4-gram     \\
 \hline
 \hline
ground truth 	& 0.878	&0.949	&0.915	&0.876\\
 \hline
CVAE 	& 0.841	&0.931	&0.900	&0.859\\
Controlled CVAE 	& 0.834	&0.927	&0.900	&0.859\\
\hline
FVN-ED 	& 0.839	&0.924	&0.898	&0.858\\
FVN-VQ 	& \textbf{0.855}	&\textbf{0.943}	&0.91	&0.869\\%& \textbf{0.860}	&0.933	&0.903	&0.863\\
%FVN-EVQ & 0.836	&0.933	&0.912	&0.874\\	%& 0.833	&0.938	&0.908	&0.868\\
%FVN 	& 0.842	&0.933	&0.903	&0.864\\
%FVN 	& \textbf{0.859}	&\textbf{0.945}	&\textbf{0.915}	&\textbf{0.876}\\%& 0.839	&\textbf{0.941}	&\textbf{0.918}	&\textbf{0.885}
%\hline 
FVN-EVQ 	& \textbf{0.855}	&\textbf{0.943}	&\textbf{0.914}	&0.876 	\\
FVN 	& 0.841	&0.935	&0.913	&\textbf{0.878} 	\\
  \end{tabular}
  }
  \caption{Diversity Evaluation on E2E. Distinct n-grams between generated texts and ground truth.}
  \label{tab:diversity}
\end{table}

We evaluate the quality and diversity of the generated text on both dataset. 
PersonageNLG is style-annotated and delexicalized, so we also report style and content correctness for it.

To evaluate quality in the generated text, we use the automatic evaluation from the E2E generation challenge, which reports BLEU (n-gram precision) \cite{papineni2002bleu}, NIST (weighted n-gram precision) \cite{doddington2002automatic}, METEOR (n-grams with synonym recall) \cite{banerjee2005meteor}, and ROUGE (n-gram recall) \cite{lin2004rouge} scores using up to 9-grams. %We report measures computed on both single ground-truth reference and multiple ground-truth references, because both sampling based methods' performance (VAEs and FVN) is impacted by the sampling.
To evaluate content correctness, we report micro precision, recall, and $F_1$ score of slot special tokens in the generated text, with respect to the slots in the given CMR $c$. 
To evaluate diversity, we report the distinct n-grams of ground-truth and baselines' examples.
For style evaluation, we separately train a personality classifier (with GloVe embeddings, 3 bi-directional LSTM layers, 2 feed-forward linear layers) on the PersonageNLG training data.
The macro precision, recall, and $F_1$ score of the personality classifier on the test set is 0.996.
We use this classifier to evaluate the style of the generated text and report our results in Table~\ref{tab:style}.

\subsection{PersonageNLG Human Evaluation}
\label{human-eval}

\begin{table}
 %\small
  \centering
  \resizebox{\columnwidth}{!}{
  \begin{tabular}{r|c|c|c}
 \multicolumn{4}{c}{}\\
    Personality           &GT      &FVN         &$p$     \\
    \hline
    agreeable           &\textbf{2.8309}         &2.4412         &***\\
    conscientiousness            &2.9808         &\textbf{2.9976}         &**\\
    disagreeable             &2.8345         &\textbf{2.9388}         &***\\
    extravert             &2.9221         &2.8933         &\\
    unconscientiousness             &\textbf{2.9365}         &2.7962         &***\\
    \hline
    overall             &\textbf{2.9001}         &2.8134         &***\\
    \multicolumn{4}{l}{*:$p<0.05$, **:$p<0.01$, ***:$p<0.001$}
  \end{tabular}
  }
  \caption{The analysis result of Question A - grammaticality / naturalness.}
  \label{tab:human_eval_a}
\end{table}

\begin{table}
  \centering
  \resizebox{0.9\columnwidth}{!}{
  \begin{tabular}{r|ccc|c}
Personality         & GT             & equal & FVN            & equal or FVN            \\ 
\hline
agreeable           & 24.46          & 22.30 & \textbf{53.24} & \textbf{\underline{75.5}4}  \\
conscientiousness   & 23.38          & 9.71  & \textbf{66.91} & \textbf{\underline{76.62}}  \\
disagreeable        & \textbf{61.87} & 12.95 & 25.18          & 38.13                   \\
extravert           & \textbf{70.14} & 9.35  & 20.50          & 29.86                   \\
unconscientiousness & \textbf{68.71} & 4.32  & 26.98          & 31.29                   \\ 
\hline
overall             & \textbf{49.71} & 11.73 & 38.56          & \textbf{\underline{50.29}} 
\end{tabular}
  }
  \caption{The analysis result of Question B - personality.  The percentage frequency distribution (\%) over three possible answers (GT, equal, FVN) for each personality is reported, with an additional column reporting the sum of equal and FVN.  In this column, underlined values are those that exceed the ones reported in the GT column.}
  \label{tab:human_eval_b}
\end{table}

In addition to automatic evaluation, we conducted a crowdsourced evaluation to compare our model against the ground truth on the entire test set.
We did not compare our model with baselines since a pilot evaluation on a random sample of 100 data points from the test set suggested that baselines did not produce fluent enough text to compare with FVN.
We considered the ground truth to be a performance upper bound and compared against it to find how close FVN is to it.
Crowdworkers were presented with a personality and two sentences (one is ground truth and the other one was generated by FVN) in random order, and were asked evaluate A) the fluency of the sentences in a scale from 1 to 3 and B) which of the two sentences was most likely to be uttered by a person with a given personality (more details in Appendix~\ref{app:human-eval-details}).
This evaluation was conducted on the entire test set consisting of 1,390 data points, 278 per personality, and each data point was judged by three different crowdworkers.

We report the result of Question A in Table~\ref{tab:human_eval_a}.
For each sentence, we averaged the scores across three judges.
The overall performance of FVN is very close to the ground truth (2.81 vs. 2.9), which suggests that FVN can generate text of comparable fluency with respect to ground truth texts.
    
We evaluated Question B using a majority vote of the three crowdworkers.
Considering the overall performance, 50.29\% of times human evaluators considered FVN generated text equal or better at conveying personality than the ground truth.
This suggests that FVN can generate text with comparable conveyance with respect to ground truth.

More details and a full breakdown on the human evaluation are available in Appendix~\ref{app:human-eval-details}.
%and in Tables~\ref{tab:human_eval_a} and ~\ref{tab:human_eval_b}.

%The most frequent codes of each slot-value are unrepeatable like the style codes in PersonageNLG dataset.
%For example the top codes for PriceRange[high] contain the linguistic patterns `high price', `expensive', `a price range of more than'.

%We train FVN on the training set and validate on the development set.
%The results reported are on the test set.

\subsection{Results and Analysis}

\begin{table*}
  \centering
  \resizebox{0.9\textwidth}{!}{
  \begin{tabular}{r|l}
  \multicolumn{2}{l}{}\\
    extravert 
    & Name EatType Food PriceRange CustomerRating Area FamilyFriendly Near \\
    \hline
    \hline
    \multirow{3}{*}{same $e^C_k$}     
    & {\color{red}\underline{let 's see what we can find on}} Name\_SLOT . {\color{red}\underline{yeah,}} it is FamilyFriendly\_SLOT with a CustomerRating\_SLOT\\ 
    &rating , it is a EatType\_SLOT , it is a Food\_SLOT place in Area\_SLOT , it is pricerange\_SLOT near Near\_SLOT .\\
    \cline{2-2}
    \multirow{3}{*}{different $e^S_n$}
    &{\color{red}\underline{i do n't know .}} Name\_SLOT is a EatType\_SLOT with a CustomerRating\_SLOT rating , also it is a FamilyFriendly\_SLOT\\
    &, Area\_SLOT , and it is a Food\_SLOT place near Near\_SLOT , also it has a price range of pricerange\_SLOT . \\
    \cline{2-2}
    &Name\_SLOT is a EatType\_SLOT , it is a FamilyFriendly\_SLOT , it 's a Food\_SLOT place , it is near Near\_SLOT ,\\
    &it has a CustomerRating\_SLOT rating , {\color{red}\underline{you know pal!}} it is in Area\_SLOT and has a price range of pricerange\_SLOT .\\\hline\hline
    \multirow{3}{*}{different $e^C_k$}
    &Name\_SLOT is a EatType\_SLOT with a CustomerRating\_SLOT rating , also it is a Food\_SLOT place , {\color{red}\underline{you know !}} \\
    &and it is Area\_SLOT , also it is FamilyFriendly\_SLOT near Near\_SLOT , also it has a price range of pricerange\_SLOT .\\
    \cline{2-2}
    \multirow{3}{*}{same $e^S_n$}
    &Name\_SLOT is a EatType\_SLOT , it is a FamilyFriendly\_SLOT , it 's a Food\_SLOT place , it is near Near\_SLOT , \\
    &it has a CustomerRating\_SLOT rating , {\color{red}\underline{{}you know}} and it is in Area\_SLOT and pricerange\_SLOT .\\
    \cline{2-2}
    &Name\_SLOT is a EatType\_SLOT , it is a Food\_SLOT place , it is FamilyFriendly\_SLOT , it 's in Area\_SLOT ,\\
    &it is near Near\_SLOT , it has a CustomerRating\_SLOT rating and a price range of pricerange\_SLOT, {\color{red}\underline{you know!}} . \\
  \end{tabular}
  }
  \caption{Diversity in FVN-generated PersonageNLG examples. Given the CMR and style the the generated text varies depending on the vector sampled from the codebook.
  }
  \label{tab:gen_variation}
\end{table*}

Tables~\ref{tab:quality}, \ref{tab:content_correctness} and \ref{tab:style} show the results on text quality, content correctness, and style. 
As shown in Table~\ref{tab:quality}, FVN significantly outperforms the state-of-the-art methods (context-m), especially on BLEU and NIST, which evaluate the precision of generated text, with the caveat regarding single or multiple references explained above.
We believe this is due to the fact that FVN explicitly models CMR and style, while context-m depends on human-engineered features.
%The single reference comparison does not reflect the fluency of FVN correctly because of inherent limitations of n-gram based measures and because FVN generates different texts depending on sampling: one generated example may not have a high n-gram overlap with a single reference but may still be fluent and correct.
%This is also the reason why scores are substantially better on the multiple references case.
Comparing FVN with CVAE and controlled CVAE, which are similar methods that also sample from the latent space, FVN performs better on all the metrics.
Human evaluation results in Section~\ref{human-eval} show that FVN is close to the ground truth in fluency and style.

Regarding the content correctness evaluation in Table~\ref{tab:content_correctness}, FVN overall performs much better than other baselines, especially on the recall score.
Methods with explicit control decoders (controlled CVAE and FVN) perform better than CVAE and FVN-EVQ, which suggests that the controlling module is useful to enhance the content conveyance.
Regarding the style evaluation in Table~\ref{tab:style}, all methods have good performance.
Style is likely easy to convey in the text (the markers are pretty specific) and easy to identify for the separately trained personality classifier. Nevertheless, FVN is the best performing model.
The text diversity comparison in Table~\ref{tab:diversity} shows how FVN and its ablations have a diversity of generated texts with respect to the ground truth texts, but so do VAE-based methods.
The combination of these findings suggests that FVN can produce text with comparable or better diversity than VAEs and ground truth, while conveying content and style more accurately.

Comparing with the ablations, the full FVN always performs better than FVN-ED and FVN-VQ, especially on the recall of slot tokens.
FVN-VQ is able to precisely generate slot tokens from the CMR, but it cannot generate all required slot tokens, while FVN can generate them with high precision and substantially higher recall.
An explanation is that the latent vectors in the content codebook only memorize the representations of texts without generalizing properly to new CMRs: since FVN is able to generate text containing most of the required slots, that text is usually longer than FVN-VQ's, which also explains why FVN performs better than FVN-VQ on METEOR and ROUGE-L that evaluate the recall of n-grams, and suggests that all encoders and codebooks are indeed needed for obtaining high performance.

The comparison between FVN and FVN-EVQ shows how in some cases FVN-EVQ has higher quality, but FVN obtains better scores on correctness and style, suggesting the additional decoder improves conveyance sacrificing some fluency.

In Table~\ref{tab:e2e}, we compare our proposed model and variants against the best performing models in the E2E challenge: TGEN~\cite{novikova2017e2e}, SLUG~\cite{juraska2018deep}, and Thomson Reuters NLG~\cite{davoodi2018e2e, smiley2018e2e}.
We can see from the results that FVN performs better than all these state-of-the-art models.
The reason of the low performance of CVAE-based methods on the E2E dataset is that the CMR are disjoint in the train and test sets (while in PersonageNLG they are overlapping) and CVAEs struggle to handle unseen CMRs.
FVNs performs well because it builds focused variations for each attribute independently instead of the entire CMR.

Table~\ref{tab:gen_variation} shows texts generated by FVN under the same CMR (given 8 attributes, rare in training data) and extravert style.
The first three samples have the same CMR latent vector, but different sampled style latent vectors.
The remaining three examples have different sampled CMR latent vectors, but the same style latent vector.
In the first three examples, the generated texts and the words representing the extravert style are different (``let 's see what we can'', ``I don't know'', ``you know'').
In the latter three examples, the words representing style are similar (``you know''), but the aggregation of attributes is different.
These examples suggest that the two codebooks learn disjoint information and that the sampling mechanism introduces the desired variation in the generated texts.

Table~\ref{tab:gen_variation} shows that FVN learns disjoint content and style codebooks and that the vectors in the codebook can be explicitly interpreted by sampling multiple texts and observing the generated patterns.
This is useful because, beyond sampling correct style vectors, we can select the realization of a style we prefer
%for conditioning the text generation 
(Table~\ref{tab:code_pattern} shows linguistic patterns associated with the top codes of each style).
These patterns are automatically learnt and suggest that there is no need to encode them with manual features.
Conditional VAEs do not provide this capability.

Samples obtained providing the same CMR and style to different models and examples of the linguistic patterns learned by FVN's style codebook are provided in Appendix~\ref{generated_samples_and_linguistic_patterns}.
Diverse samples obtained from FVN by sampling different latent codes are shown in Table~\ref{tab:e2e_diversity}.

\section{Conclusion}
%In this paper, we studied the style-variation text generation problem. 
%We propose a novel focal variation network (FVN), which is based on VQ-VAE that models the focal variation of the given theme over the full latent space.
%Experimental results show that FVN is promising and achieves state-of-the-art performance on the PersonageNLG dataset. 
%Further, we analyze the performance of combining disentangled latent vectors for text generation.
%We discuss how the model learns disentangled latent spaces for different dimensions of variation (meaning representation and style) and is able to combine them to generate texts that reflect those dimensions for variation accurately.

In this paper, we studied the task of controlling language generation, with a specific emphasis on content conveyance and style variation.
We introduced FVN, a novel model that overcomes the limitations of previous models, namely lack of conveyance and lack of diversity of the generated text, by adopting disjoint discrete latent spaces for each of the desired attributes.
%and extending the basic VQ-VAE.
Our experimental results show that FVN achieves state-of-the-art performance on PersonageNLG and E2E datasets and generated texts are comparable to ground truth ones according to human evaluators.

\bibliography{anthology,emnlp2020}

\begin{thebibliography}{47}
\expandafter\ifx\csname natexlab\endcsname\relax\def\natexlab#1{#1}\fi

\bibitem[{Bahdanau et~al.(2015)Bahdanau, Cho, and Bengio}]{bahdanau2014neural}
Dzmitry Bahdanau, Kyunghyun Cho, and Yoshua Bengio. 2015.
\newblock Neural machine translation by jointly learning to align and
  translate.
\newblock In \emph{International Conference on Learning Representations, San
  Diego, California, USA}.

\bibitem[{Banerjee and Lavie(2005)}]{banerjee2005meteor}
Satanjeev Banerjee and Alon Lavie. 2005.
\newblock Meteor: An automatic metric for mt evaluation with improved
  correlation with human judgments.
\newblock In \emph{Proceedings of the acl workshop on intrinsic and extrinsic
  evaluation measures for machine translation and/or summarization}, pages
  65--72.

\bibitem[{Biber(1991)}]{biber1991variation}
Douglas Biber. 1991.
\newblock \emph{Variation across speech and writing}.
\newblock Cambridge University Press.

\bibitem[{Bird et~al.(2009)Bird, Klein, and Loper}]{Bird2009NLP}
Steven Bird, Ewan Klein, and Edward Loper. 2009.
\newblock \emph{Natural Language Processing with Python}.
\newblock O'Reilly Media, Inc.

\bibitem[{Bowman et~al.(2016)Bowman, Vilnis, Vinyals, Dai, Jozefowicz, and
  Bengio}]{bowman2016generating}
Samuel Bowman, Luke Vilnis, Oriol Vinyals, Andrew Dai, Rafal Jozefowicz, and
  Samy Bengio. 2016.
\newblock Generating sentences from a continuous space.
\newblock In \emph{Proceedings of The 20th SIGNLL Conference on Computational
  Natural Language Learning}, pages 10--21.

\bibitem[{Dai et~al.(2019)Dai, Liang, Qiu, and Huang}]{dai2019transformer}
Ning Dai, Jianze Liang, Xipeng Qiu, and Xuanjing Huang. 2019.
\newblock \href {http://arxiv.org/abs/1905.05621} {Style transformer: Unpaired
  text style transfer without disentangled latent representation}.
\newblock \emph{CoRR}, abs/1905.05621.

\bibitem[{Dathathri et~al.(2020)Dathathri, Madotto, Lan, Hung, Frank, Molino,
  Yosinski, and Liu}]{Dathathri2020PlugAP}
Sumanth Dathathri, Andrea Madotto, Janice Lan, Jane Hung, Eric~C. Frank, Piero
  Molino, Jason Yosinski, and Rosanne Liu. 2020.
\newblock Plug and play language models: A simple approach to controlled text
  generation.
\newblock \emph{ArXiv}, abs/1912.02164.

\bibitem[{Davoodi et~al.(2018)Davoodi, Smiley, Song, and
  Schilder}]{davoodi2018e2e}
Elnaz Davoodi, Charese Smiley, Dezhao Song, and Frank Schilder. 2018.
\newblock The e2e nlg challenge: Training a sequence-to-sequence approach for
  meaning representation to natural language sentences.
\newblock In \emph{in prep. for INLG conference}.

\bibitem[{Doddington(2002)}]{doddington2002automatic}
George Doddington. 2002.
\newblock Automatic evaluation of machine translation quality using n-gram
  co-occurrence statistics.
\newblock In \emph{Proceedings of the second international conference on Human
  Language Technology Research}, pages 138--145.

\bibitem[{Du{\v{s}}ek and Jurcicek(2016{\natexlab{a}})}]{duvsek2016context}
Ond{\v{r}}ej Du{\v{s}}ek and Filip Jurcicek. 2016{\natexlab{a}}.
\newblock A context-aware natural language generator for dialogue systems.
\newblock In \emph{Proceedings of the 17th Annual Meeting of the Special
  Interest Group on Discourse and Dialogue}, pages 185--190.

\bibitem[{Du{\v{s}}ek and Jurcicek(2016{\natexlab{b}})}]{duvsek2016sequence}
Ond{\v{r}}ej Du{\v{s}}ek and Filip Jurcicek. 2016{\natexlab{b}}.
\newblock Sequence-to-sequence generation for spoken dialogue via deep syntax
  trees and strings.
\newblock In \emph{Proceedings of the 54th Annual Meeting of the Association
  for Computational Linguistics (Volume 2: Short Papers)}, pages 45--51.

\bibitem[{Du{\v{s}}ek et~al.(2018)Du{\v{s}}ek, Novikova, and
  Rieser}]{duvsek2018findings}
Ond{\v{r}}ej Du{\v{s}}ek, Jekaterina Novikova, and Verena Rieser. 2018.
\newblock Findings of the e2e nlg challenge.
\newblock In \emph{Proceedings of the 11th International Conference on Natural
  Language Generation}, pages 322--328.

\bibitem[{Du{\v{s}}ek et~al.(2020)Du{\v{s}}ek, Novikova, and
  Rieser}]{dusek.etal2020:csl}
Ond\v{r}ej Du{\v{s}}ek, Jekaterina Novikova, and Verena Rieser. 2020.
\newblock \href {https://doi.org/10.1016/j.csl.2019.06.009} {Evaluating the
  {{State}}-of-the-{{Art}} of {{End}}-to-{{End Natural Language Generation}}:
  {{The E2E NLG Challenge}}}.
\newblock \emph{Computer Speech \& Language}, 59:123--156.

\bibitem[{Ficler and Goldberg(2017)}]{ficler2017controlling}
Jessica Ficler and Yoav Goldberg. 2017.
\newblock Controlling linguistic style aspects in neural language generation.
\newblock In \emph{Proceedings of the Workshop on Stylistic Variation}, pages
  94--104.

\bibitem[{Harrison et~al.(2019)Harrison, Reed, Oraby, and
  Walker}]{harrison2019maximizing}
Vrindavan Harrison, Lena Reed, Shereen Oraby, and Marilyn Walker. 2019.
\newblock Maximizing stylistic control and semantic accuracy in nlg:
  Personality variation and discourse contrast.
\newblock \emph{DSNNLG 2019}, page~1.

\bibitem[{Holtzman et~al.(2020)Holtzman, Buys, Forbes, and
  Choi}]{Holtzman2020TheCC}
Ari Holtzman, Jan Buys, Maxwell Forbes, and Yejin Choi. 2020.
\newblock The curious case of neural text degeneration.
\newblock \emph{ArXiv}, abs/1904.09751.

\bibitem[{Hu et~al.(2017)Hu, Yang, Liang, Salakhutdinov, and
  Xing}]{hu2017toward}
Zhiting Hu, Zichao Yang, Xiaodan Liang, Ruslan Salakhutdinov, and Eric~P Xing.
  2017.
\newblock Toward controlled generation of text.
\newblock In \emph{Proceedings of the 34th International Conference on Machine
  Learning-Volume 70}, pages 1587--1596.

\bibitem[{Juraska et~al.(2018)Juraska, Karagiannis, Bowden, and
  Walker}]{juraska2018deep}
Juraj Juraska, Panagiotis Karagiannis, Kevin Bowden, and Marilyn Walker. 2018.
\newblock A deep ensemble model with slot alignment for sequence-to-sequence
  natural language generation.
\newblock In \emph{Proceedings of the 2018 Conference of the North American
  Chapter of the Association for Computational Linguistics: Human Language
  Technologies, Volume 1 (Long Papers)}, pages 152--162.

\bibitem[{Juraska and Walker(2018)}]{juraska2018characterizing}
Juraj Juraska and Marilyn Walker. 2018.
\newblock Characterizing variation in crowd-sourced data for training neural
  language generators to produce stylistically varied outputs.
\newblock In \emph{Proceedings of the 11th International Conference on Natural
  Language Generation}, pages 441--450.

\bibitem[{Khandelwal et~al.(2020)Khandelwal, Levy, Jurafsky, Zettlemoyer, and
  Lewis}]{Khandelwal2020GeneralizationTM}
Urvashi Khandelwal, Omer Levy, Dan Jurafsky, Luke Zettlemoyer, and Mike Lewis.
  2020.
\newblock Generalization through memorization: Nearest neighbor language
  models.
\newblock \emph{ArXiv}, abs/1911.00172.

\bibitem[{Kikuchi et~al.(2014)Kikuchi, Hirao, Takamura, Okumura, and
  Nagata}]{kikuchi-etal-2014-single}
Yuta Kikuchi, Tsutomu Hirao, Hiroya Takamura, Manabu Okumura, and Masaaki
  Nagata. 2014.
\newblock \href {https://doi.org/10.3115/v1/P14-2052} {Single document
  summarization based on nested tree structure}.
\newblock In \emph{Proceedings of the 52nd Annual Meeting of the Association
  for Computational Linguistics (Volume 2: Short Papers)}, pages 315--320,
  Baltimore, Maryland. Association for Computational Linguistics.

\bibitem[{Kingma and Ba(2015)}]{kingma2014adam}
Diederik~P Kingma and Jimmy Ba. 2015.
\newblock Adam: A method for stochastic optimization.
\newblock In \emph{International Conference on Learning Representations, San
  Diego, California, USA}.

\bibitem[{Klein et~al.(2017)Klein, Kim, Deng, Senellart, and Rush}]{opennmt}
Guillaume Klein, Yoon Kim, Yuntian Deng, Jean Senellart, and Alexander~M. Rush.
  2017.
\newblock \href {https://doi.org/10.18653/v1/P17-4012} {Open{NMT}: Open-source
  toolkit for neural machine translation}.
\newblock In \emph{Proc. ACL}.

\bibitem[{Lin(2004)}]{lin2004rouge}
Chin-Yew Lin. 2004.
\newblock \href {https://www.aclweb.org/anthology/W04-1013} {{ROUGE}: A package
  for automatic evaluation of summaries}.
\newblock In \emph{Text Summarization Branches Out}, pages 74--81, Barcelona,
  Spain. Association for Computational Linguistics.

\bibitem[{Mairesse and Walker(2007)}]{mairesse2007personage}
Fran{\c{c}}ois Mairesse and Marilyn Walker. 2007.
\newblock Personage: Personality generation for dialogue.
\newblock In \emph{Proceedings of the 45th Annual Meeting of the Association of
  Computational Linguistics}, pages 496--503.

\bibitem[{Novikova et~al.(2017)Novikova, Du{\v{s}}ek, and
  Rieser}]{novikova2017e2e}
Jekaterina Novikova, Ond{\v{r}}ej Du{\v{s}}ek, and Verena Rieser. 2017.
\newblock The e2e dataset: New challenges for end-to-end generation.
\newblock In \emph{Proceedings of the 18th Annual SIGdial Meeting on Discourse
  and Dialogue}, pages 201--206.

\bibitem[{Novikova et~al.(2016)Novikova, Lemon, and Rieser}]{novikova2016crowd}
Jekaterina Novikova, Oliver Lemon, and Verena Rieser. 2016.
\newblock Crowd-sourcing nlg data: Pictures elicit better data.
\newblock In \emph{Proceedings of the 9th International Natural Language
  Generation conference}, pages 265--273.

\bibitem[{van~den Oord et~al.(2017)van~den Oord, Vinyals
  et~al.}]{van2017neural}
Aaron van~den Oord, Oriol Vinyals, et~al. 2017.
\newblock Neural discrete representation learning.
\newblock In \emph{Advances in Neural Information Processing Systems}, pages
  6306--6315.

\bibitem[{Oraby et~al.(2019)Oraby, Harrison, Ebrahimi, and
  Walker}]{oraby2019curate}
Shereen Oraby, Vrindavan Harrison, Abteen Ebrahimi, and Marilyn Walker. 2019.
\newblock Curate and generate: A corpus and method for joint control of
  semantics and style in neural nlg.
\newblock In \emph{Proceedings of the 57th Annual Meeting of the Association
  for Computational Linguistics}, pages 5938--5951.

\bibitem[{Oraby et~al.(2018)Oraby, Reed, Tandon, Sharath, Lukin, and
  Walker}]{oraby2018controlling}
Shereen Oraby, Lena Reed, Shubhangi Tandon, TS~Sharath, Stephanie Lukin, and
  Marilyn Walker. 2018.
\newblock Controlling personality-based stylistic variation with neural natural
  language generators.
\newblock In \emph{Proceedings of the 19th Annual SIGdial Meeting on Discourse
  and Dialogue}, pages 180--190.

\bibitem[{Paiva and Evans(2004)}]{paiva2004framework}
Daniel~S Paiva and Roger Evans. 2004.
\newblock A framework for stylistically controlled generation.
\newblock In \emph{International Conference on Natural Language Generation},
  pages 120--129. Springer.

\bibitem[{Papineni et~al.(2002)Papineni, Roukos, Ward, and
  Zhu}]{papineni2002bleu}
Kishore Papineni, Salim Roukos, Todd Ward, and Wei{-}Jing Zhu. 2002.
\newblock Bleu: a method for automatic evaluation of machine translation.
\newblock In \emph{{ACL}}, pages 311--318. {ACL}.

\bibitem[{Pennington et~al.(2014)Pennington, Socher, and
  Manning}]{pennington2014glove}
Jeffrey Pennington, Richard Socher, and Christopher~D. Manning. 2014.
\newblock Glove: Global vectors for word representation.
\newblock In \emph{{EMNLP}}, pages 1532--1543. {ACL}.

\bibitem[{Radford et~al.(2018)Radford, Narasimhan, Salimans, and
  Sutskever}]{radford2018improving}
Alec Radford, Karthik Narasimhan, Tim Salimans, and Ilya Sutskever. 2018.
\newblock Improving language understanding by generative pre-training.
\newblock \emph{URL https://s3-us-west-2. amazonaws.
  com/openai-assets/researchcovers/languageunsupervised/language understanding
  paper. pdf}.

\bibitem[{Radford et~al.(2019)Radford, Wu, Child, Luan, Amodei, and
  Sutskever}]{radford2019language}
Alec Radford, Jeffrey Wu, Rewon Child, David Luan, Dario Amodei, and Ilya
  Sutskever. 2019.
\newblock Language models are unsupervised multitask learners.
\newblock \emph{OpenAI Blog}, 1(8).

\bibitem[{Rashkin et~al.(2020)Rashkin, Celikyilmaz, Choi, and
  Gao}]{rashkin2020plotmachines}
Hannah Rashkin, Asli Celikyilmaz, Yejin Choi, and Jianfeng Gao. 2020.
\newblock Plotmachines: Outline-conditioned generation with dynamic plot state
  tracking.
\newblock \emph{arXiv preprint arXiv:2004.14967}.

\bibitem[{Rashkin et~al.(2018)Rashkin, Smith, Li, and
  Boureau}]{rashkin2018towards}
Hannah Rashkin, Eric~Michael Smith, Margaret Li, and Y-Lan Boureau. 2018.
\newblock Towards empathetic open-domain conversation models: A new benchmark
  and dataset.
\newblock \emph{arXiv preprint arXiv:1811.00207}.

\bibitem[{dos Santos et~al.(2018)dos Santos, Melnyk, and
  Padhi}]{dossantos18fightingoffensivelanguage}
C{\'{\i}}cero~Nogueira dos Santos, Igor Melnyk, and Inkit Padhi. 2018.
\newblock \href {https://doi.org/10.18653/v1/P18-2031} {Fighting offensive
  language on social media with unsupervised text style transfer}.
\newblock In \emph{Proceedings of the 56th Annual Meeting of the Association
  for Computational Linguistics, {ACL} 2018, Melbourne, Australia, July 15-20,
  2018, Volume 2: Short Papers}, pages 189--194. Association for Computational
  Linguistics.

\bibitem[{Shen et~al.(2018)Shen, Su, Niu, and Demberg}]{shen2018improving}
Xiaoyu Shen, Hui Su, Shuzi Niu, and Vera Demberg. 2018.
\newblock Improving variational encoder-decoders in dialogue generation.
\newblock In \emph{Thirty-Second AAAI Conference on Artificial Intelligence}.

\bibitem[{{Shirish Keskar} et~al.(2019){Shirish Keskar}, {McCann}, {Varshney},
  {Xiong}, and {Socher}}]{keskar2019ctrl}
Nitish {Shirish Keskar}, Bryan {McCann}, Lav~R. {Varshney}, Caiming {Xiong},
  and Richard {Socher}. 2019.
\newblock \href {http://arxiv.org/abs/1909.05858} {{CTRL: A Conditional
  Transformer Language Model for Controllable Generation}}.
\newblock \emph{arXiv e-prints}, page arXiv:1909.05858.

\bibitem[{Smiley et~al.(2018)Smiley, Davoodi, Song, and
  Schilder}]{smiley2018e2e}
Charese Smiley, Elnaz Davoodi, Dezhao Song, and Frank Schilder. 2018.
\newblock The e2e nlg challenge: End-to-end generation through partial template
  mining.
\newblock \emph{in prep}.

\bibitem[{Sohn et~al.(2015)Sohn, Lee, and Yan}]{sohn2015learning}
Kihyuk Sohn, Honglak Lee, and Xinchen Yan. 2015.
\newblock Learning structured output representation using deep conditional
  generative models.
\newblock In \emph{Advances in neural information processing systems}, pages
  3483--3491.

\bibitem[{Song et~al.(2019)Song, Zhang, Cui, Wang, and
  Liu}]{song2019exploiting}
Haoyu Song, Wei-Nan Zhang, Yiming Cui, Dong Wang, and Ting Liu. 2019.
\newblock Exploiting persona information for diverse generation of
  conversational responses.
\newblock \emph{arXiv preprint arXiv:1905.12188}.

\bibitem[{Wang et~al.(2018)Wang, Wang, Tur, and Williams}]{Wang18NipsConvAI}
Yi-Chia Wang, Runze Wang, Gokhan Tur, and Hugh Williams. 2018.
\newblock Can you be more polite and positive? infusing social language into
  task-oriented conversational agents.
\newblock In \emph{NeurIPS 2018 Workshop on the Second Conversational AI,}.

\bibitem[{Ye et~al.(2020)Ye, Shi, Zhou, Wei, and Li}]{ye2020variational}
Rong Ye, Wenxian Shi, Hao Zhou, Zhongyu Wei, and Lei Li. 2020.
\newblock Variational template machine for data-to-text generation.
\newblock \emph{arXiv preprint arXiv:2002.01127}.

\bibitem[{Zhang et~al.(2019)Zhang, Wang, Zhang, Zhang, and
  Gai}]{zhang2019improve}
Yuchi Zhang, Yongliang Wang, Liping Zhang, Zhiqiang Zhang, and Kun Gai. 2019.
\newblock Improve diverse text generation by self labeling conditional
  variational auto encoder.
\newblock In \emph{ICASSP 2019-2019 IEEE International Conference on Acoustics,
  Speech and Signal Processing (ICASSP)}, pages 2767--2771. IEEE.

\bibitem[{Ziegler et~al.(2019)Ziegler, Stiennon, Wu, Brown, Radford, Amodei,
  Christiano, and Irving}]{ziegler2019finetuning}
Daniel~M. Ziegler, Nisan Stiennon, Jeffrey Wu, Tom~B. Brown, Alec Radford,
  Dario Amodei, Paul Christiano, and Geoffrey Irving. 2019.
\newblock \href {https://arxiv.org/abs/1909.08593} {Fine-tuning language models
  from human preferences}.
\newblock \emph{arXiv preprint arXiv:1909.08593}.

\end{thebibliography}
\bibliographystyle{acl_natbib}

\clearpage
\appendix

\section{Implementation details}
\label{app:implementation_details}

\begin{table}
  \centering
  \resizebox{\columnwidth}{!}{
  \begin{tabular}{r|l}
Module & Layers (in, out)\\\hline
content codebook & Emb(K, D)\\
style/slot-value codebook & Emb(N, D)\\\hline
text-to-content encoder & Emb($|V|$, D), Bi-LSTM(D, D)\\
text-to-style encoder & Emb($|V|$, D), Bi-LSTM(D, D) \\\hline
content encoder & Emb($|V|$, D), Bi-LSTM(D, D)\\
style encoder & Emb($|V|$, D), Bi-LSTM(D, D)\\\hline
content decoder & Dense(D, D/2), Dense(D/2, 8)\\
style decoder & Dense(D, D/2), Dense(D/2, 5)\\
slot-value decoder & Dense(D, D/2), Dense(D/2, 36)\\\hline
\multirow{2}{*}{text decoder} & Emb($|V|$, D), LSTM(D,2D), Attn(2D, 2D) \\
                & Dense(2D, D), Dense(D,$|V|$) \\

  \end{tabular}
  }
  \caption{Details of Modules in FVN: $D = 300$, $K = 512$, $N = |V|$}
  \label{tab:modules}
\end{table}

We use NLTK~\cite{Bird2009NLP} to tokenize each sentence and de-lexicalize the text as described in \cite{duvsek2016context}.
We use 300-dimensional GloVe embeddings~\cite{pennington2014glove} trained on 840B words.
Words not in GloVe are initialized as the averaged embeddings of all other embeddings plus a small amount of random noise to make them different from each other. 
The details of each module in the FVN are listed in Table \ref{tab:modules}. 
We set $D = 300$, $K = 512$, $N = |V|$. The encoders are three-layer stacked Bi-LSTM and the text decoder is one-layer LSTM. The style/slot-value codebook is initialized as pre-trained word embedding.
The content codebook is uniformed initialized in the range of $[-1/K, 1/K]$.
We use the Adam optimizer~\cite{kingma2014adam} with a learning rate of 0.001 for minimizing the total loss.

\section{Experiments details}
\label{app:experiment-details}

These are the links to the adopted datasets and the code for computing the metrics.

PersonageNLG text generation dataset: \url{https://nlds.soe.ucsc.edu/stylistic-variation-nlg}

End-2-End Challenge dataset (E2E): \url{http://www.macs.hw.ac.uk/InteractionLab/E2E/}

Automatic evaluation metrics code from the E2E generation challenge: \url{https://github.com/tuetschek/e2e-metrics}

Distinct n-grams metric code: \url{https://github.com/neural-dialogue-metrics}

Table~\ref{tab:dataseTCr} shows details of the PersonageNLG dataset, while Table~\ref{tab:dataseTCr_e2e} shows details of the E2E dataset.

\begin{table}
  \centering
  \resizebox{\columnwidth}{!}{
  \begin{tabular}{r||c||c||c|c|c|c|c|c}
Dataset & Pairs &\multicolumn{6}{|c}{Number of Slots in CMR}\\
\hline
        &       &CMRs &3      &4      &5      &6      &7      &8\\
\hline
Train   &88,855  &600 &0.13   &0.30   &0.29   &0.22   &0.06   &0.01\\
Test    &1,390   &35 &0.02   &0.04   &0.06   &0.15   &0.35   &0.37\\
  \end{tabular}
  }
  \caption{Distribution of slots in the CMR in both training and test splits of PersonageNLG. Pairs refer to content-utterance pairs.}
  \label{tab:dataseTCr}
\end{table}

\begin{table}
  \centering
  \resizebox{\columnwidth}{!}{
  \begin{tabular}{r||c||c||c|c|c|c|c|c}
Dataset & Pairs & CMRs & \multicolumn{6}{|c}{Number of Slots in CMR}\\
\hline
        &   &CMRs    &3      &4      &5      &6      &7      &8\\
\hline
Train   &42061  &4862   &0.05   &0.18   &0.32   &0.28   &0.14   &0.03\\
Dev     &4672   &547    &0.09   &0.11   &0.05   &0.35   &0.30   &0.10\\
Test    &4693   &630    &0.01   &0.03   &0.08   &0.17   &0.34   &0.37\\
  \end{tabular}
  }
  \caption{Distribution of slots in the CMR in both training, development and test splits of E2E. Pairs refer to content-utterance pairs.}
  \label{tab:dataseTCr_e2e}
\end{table}

\section{Baselines details}
\label{app:baseline-details}
The first three baselines are taken from \cite{oraby2018controlling} and adopt the TGen architecture \cite{duvsek2016sequence}, an encoder-decoder network, with different kinds of input.

%\textbf{nosupervision (NOSUP)} follows \cite{duvsek2016sequence}, which is trained using all 88K utterances in a single pool for up to 14 epochs based on loss monitoring for the decoder and re-ranker.

\textbf{TOKEN} adds a token of additional supervision to encode personality.
Unlike other works that use a single token to control the generator's output~\cite{hu2017toward}, the personality token encodes a several different parameters that define style.

\textbf{CONTEXT} introduces a context vector that explicitly encodes a set of 36 manually-defined style parameters encoded as a vector of binary values.
%and a feed-forward network is added as a context encoder, taking the vector as input to the hidden state of the encoder.
%and making the parameters available at every time step to a multiplicative attention unit. 
We then apply these style encoding approaches to three state of the art models taken from \cite{harrison2019maximizing}, which extend \cite{oraby2018controlling} by changing the basic encoder-decoder network to OpenNMT-py~\cite{opennmt} in the following ways.

\textbf{m1} inserts style information into the sequence of tokens that constitute the content $c$;

\textbf{m2} incorporates style information in the content encoding process by concatenating style representation with content representation before passing it to the content encoder;

\textbf{m3} incorporates style information into the generation process by additional inputs to the decoder. At each decoding step, style representation is concatenated with each word's embedding and passed as input to the decoder.

\textbf{token-m} means that style (personality here) is encoded with a single token; 

\textbf{context-m} means that style is encoded via the 36 parameters.

\textbf{TGEN}~\cite{novikova2017e2e} adopts a seq2seq model with attention \cite{bahdanau2014neural} with added beam search and a reranker penalizing outputs that stray away from the input CMR.

\textbf{SLUG}~\cite{juraska2018deep} adopts seq2seq-based ensemble which uses LSTM/CNN as the encoders and LSTM as the decoder); heuristic slot aligner reranking and data augmentation.
Both TGEN and SLUG use partial (`name' and `near' slot) de-lexicalized texts .

\textbf{Thomson Reuters NLG}~\cite{davoodi2018e2e, smiley2018e2e} use fully de-lexicalized text and a seq2seq model with hyperparameter tunning.

\section{Human evaluation details}
\label{app:human-eval-details}

Crowdworkers were presented with a personality and two sentences (one is ground truth and the other one was generated by our model) in random order, and were asked to answer the following two questions: 

\begin{itemize}
\item Question A: On a scale of 1-3, how grammatical or natural is this sentence? (please answer for both sentences).

\item Question B: Which of these two sentences do you think would be more likely to be said by a(n) \_\_\_ person? (Fill in \_\_\_ with the personality given, e.g. agreeable) 
Answers: Sentence 1, 2, equally

\end{itemize}

Question A asked the crowdworkers to assess the degree of grammaticality / naturalness of a sentence while Question B was designed to evaluate which of the two sentences exhibits a specific personality.

We report the result of Question A in Table~\ref{tab:human_eval_a}.
For each sentence, we averaged the scores across three judges, and conducted a paired t-test between the ground truth and our model for each personality.
The result shows that the FVN sentences were considered significantly more grammatical / natural on conscientiousness and disagreeableness, the ground truth sentences were better on agreeable and unconscientiousness, and no difference was found for extravert.
The overall performance of FVN is very close to the ground truth (2.81 vs. 2.9), which suggests that FVN can generate text of comparable fluency with respect to ground truth texts.
    
We evaluated Question B using a majority vote of the three crowdworkers.
Table~\ref{tab:human_eval_b} shows the percentage frequency distribution for each personality and the entire test set.
We found that our FVN model performs better than the ground truth on agreeable and conscientiousness, while the ground truth is better for the rest of the three personalities.
Specifically, 53\% and 67\% of the time, the crowdworkers judge the agreeable and conscientious sentences generated by our model to be better than the ground truth sentences.
This finding is surprising, since we consider the ground truth be an upper bound in this task, and our model outperforms it two out of five personalities.
One possible explanation about why FVN only performs better on agreeable and conscientiousness is that the language patterns of agreeableness and conscientiousness are more systematic and thus easier to learn by the model.
In Table~\ref{tab:human_eval_b} we also report a column that shows the percentage frequency of text where the judgment was equal or in favor of FVN.
Underlined rows show when the number of equal judgments or judgments favorable to FVN exceeds the judgments that preferred the ground truth text.
Considering the overall performance, 50.29\% of times human evaluators considered FVN generated text equal or better at conveying personality than the ground truth.
This finding suggests that FVN can generate text with comparable conveyance with respect to ground truth texts.

\begin{table*}
  \centering
  \resizebox{\textwidth}{!}{
  \begin{tabular}{r|l}
  \multicolumn{2}{l}{}\\
    agreeable & ``let 's see what we can find on''  ``well , i see''   ``did you say ?''  ``i suppose''  ``right'' ``okay ?'' `` you see ?'' ``it is somewhat''\\
    \hline
    disagreeable & ``oh god i mean , everybody knows'' ``oh god'' ``i do n't know .'' ``i am not sure .''\\
    \hline
    conscientious & ``let 's see what we can find on'' ``well , i see'' ``did you say '' `` sort of '' ``you see ?'' ``let 's see, '' ``...''\\
    \hline
    unconscientious & ``oh god i , i do n't know .'' ``darn'' `` i mean .'' ``i ... i , i do n't know .'' ``i mean , i am not sure .'' ``damn'' ``!'' ``it has like a ''\\
    \hline
    extravert & ``oh god i am not sure .'' ``let 's see ,'' ``...'' ``alright ?'' ``yeah'' ``i do n't know'' ``did you say ?" ``you know !'' ``you know and'' ``pal'' ``! ''\\
  \end{tabular}
  }
  \caption{Top codes' linguistic pattern of each style}
  \label{tab:code_pattern}
\end{table*}

\begin{table*}[ht!]
  \centering
  \resizebox{\textwidth}{!}{
  \begin{tabular}{l}
  \hline
area[city centre] customer rating[5 out of 5] eatType[pub] familyFriendly[no] food[French] name[The Phoenix] near[Crowne Plaza Hotel] priceRange[more than £30]\\\hline
-use the top frequent code of each value-\\
The Phoenix is a french pub near Crowne Plaza Hotel in the city centre . It is not children friendly and has a price range of more than £30 and has a customer rating of 5 out of 5 .\\\hline\hline

-use same area[city centre] code, other values' codes are sampled-\\
The Phoenix is a pub in the city centre . It is a french food . It is located in the city centre . \\
The Phoenix is a pub in the city centre . It is a french food . It is a high price range and is not child friendly . \\\hline

-use same customer rating[5 out of 5] code, other values' codes are sampled-\\
The Phoenix is a french pub located in the city centre . It is a high customer rating and is not children friendly .\\
The Phoenix is a pub in the city centre near Crowne Plaza Hotel . It is a high customer rating and is not children friendly . \\\hline

-use same eattype[pub] code, other values' codes are sampled- \\
The Phoenix is a french pub near Crowne Plaza Hotel in the city centre . It is not children friendly and has a price range of more than £30 . \\
The Phoenix is a french pub in the city centre near Crowne Plaza Hotel . It is not child friendly and has a high price range and a customer rating of 5 out of 5 .\\\hline

-use same familyFriendly[no] code, other values' codes are sampled- \\
The Phoenix is a french pub located in the city centre near Crowne Plaza Hotel . It is not family-friendly and has a customer rating of 5 out of 5 . \\
The Phoenix is a french pub located in the city centre . It is not family-friendly and has a customer rating of 5 out of 5 . \\\hline

-use same food[French] code, other values' codes are sampled-\\
The Phoenix is a french pub in the city centre . It is a high customer rating and is not children friendly . \\
The Phoenix is a french pub located in the city centre . It is not family-friendly . \\\hline

-use same priceRange[more than £30] code, other values' codes are sampled-\\
The Phoenix is a french pub in the city centre . It is not children friendly and has a price range of more than £30 .\\
The Phoenix is a french pub near Crowne Plaza Hotel in the city centre . It is not children friendly and has a price range of more than £30 .\\\hline

  \end{tabular}
  }
  \caption{Diversity in FVN-generated E2E examples.}
  \label{tab:e2e_diversity}
\end{table*}

\section{Generated Samples and Linguistic Patterns}
\label{generated_samples_and_linguistic_patterns}

Table~\ref{tab:e2e_diversity} shows generated examples from FVN trained on E2E.
Given a CMR, we sample a code for each slot value.
The first part shows the generated text using the most frequent code for each slot value.
We can see that the text is fluent and conveys the CMR precisely.
In the remaining part, we keep one slot-value's code fixed and the remaining slot codes are sampled.
The fixed slot-value is present, but some of the other slot-values are missing in the generated text.
One explanation is that in the training data the text associated with a CMR can also contain missing values and therefore the codebook memorizes this behavior.

\end{document}